\documentclass[conference]{IEEEtran}
\IEEEoverridecommandlockouts
\usepackage{cite}
\usepackage{amsmath,amssymb,amsfonts}
\usepackage{algorithmic}
\usepackage{graphicx}
\usepackage{textcomp}
\usepackage{xcolor}
\usepackage{float}
\usepackage[bookmarks=false,hidelinks]{hyperref}
\usepackage{pdfprivacy}
\usepackage{booktabs}
\usepackage{graphicx}
\usepackage{float}
\usepackage{placeins} 
\usepackage[margin=1in]{geometry} 
\usepackage{graphicx}
\usepackage{booktabs} 
\usepackage{array}    
\usepackage{tabularx} 
\usepackage{placeins} 
\usepackage{url}

\usepackage[a-1b]{pdfx}
\def\BibTeX{{\rm B\kern-.05em{\sc i\kern-.025em b}\kern-.08em
    T\kern-.1667em\lower.7ex\hbox{E}\kern-.125emX}}
\begin{document}

\title{Lightweight Object Detection Using Quantized YOLOv4-Tiny for Emergency Response in Aerial Imagery}

\author{
    \IEEEauthorblockN{Sindhu Boddu}
    \IEEEauthorblockA{
        \textit{Department of Electrical } \\
		\textit{and Computer Engineering} \\        
        \textit{UNC Charlotte}\\
        Charlotte, North Carolina, USA \\
        sboddu2@charlotte.edu}
    \and
    \IEEEauthorblockN{Dr. Arindam Mukherjee}
    \IEEEauthorblockA{
        \textit{Department of Electrical  } \\
        \textit{ and Computer Engineering} \\ 
        \textit{UNC Charlotte}\\
        Charlotte, North Carolina, USA \\
        amukherj@charlotte.edu}
    }

\maketitle

\begin{abstract}
This paper presents a lightweight and energy-efficient object detection solution for aerial imagery captured during emergency response situations. We focus on deploying the YOLOv4-Tiny model, a compact convolutional neural network, optimized through post-training quantization to INT8 precision. The model is trained on a custom-curated aerial emergency dataset, consisting of 10,820 annotated images covering critical emergency scenarios. Unlike prior works that rely on publicly available datasets, we created this dataset ourselves due to the lack of publicly available drone-view emergency imagery, making the dataset itself a key contribution of this work. The quantized model is evaluated against YOLOv5-small across multiple metrics, including mean Average Precision (mAP), F1 score, inference time, and model size. Experimental results demonstrate that the quantized YOLOv4-Tiny achieves comparable detection performance while reducing the model size from 22.5 MB to 6.4 MB and improving inference speed by 44\%. With a 71\% reduction in model size and a 44\% increase in inference speed, the quantized YOLOv4-Tiny model proves highly suitable for real-time emergency detection on low-power edge devices.
\end{abstract}


\section{Introduction}
The integration of artificial intelligence into embedded systems has revolutionized real-time monitoring applications, particularly in critical scenarios such as emergency response and disaster management. Drones, equipped with onboard cameras and lightweight object detection algorithms, are increasingly being used to gather visual information from aerial perspectives. These platforms are invaluable in detecting and localizing emergency incidents including traffic accidents, overturned vehicles, fire outbreaks, and the presence of emergency services such as ambulances or police units. However, achieving accurate and fast object detection in such settings remains a challenge, especially when operating on resource-constrained edge devices like the Raspberry Pi.

Traditional object detection frameworks such as Faster R-CNN and SSD are known for their high accuracy but suffer from considerable latency and computational requirements, rendering them unsuitable for deployment on low-power devices. Even modern real-time detectors like YOLOv3 and YOLOv5-small, though efficient, can benefit from additional optimizations when targeting battery-powered embedded platforms. To meet the strict requirements of real-time inference, energy efficiency, and small memory footprint, lightweight architectures must be considered alongside model compression techniques such as quantization.

In this context, we explore the deployment of a quantized version of YOLOv4-Tiny, a compact convolutional neural network optimized for object detection, specifically targeting aerial emergency imagery. The model is trained on a domain-specific dataset consisting of 10,820 annotated aerial images covering seven emergency-related classes, including police vehicles, fire engines, ambulances, car crashes, and more. \textit{Unlike prior works that rely on public datasets, this dataset was curated from scratch due to the absence of publicly available drone-view emergency imagery, making it a significant contribution of this work.} To ensure compatibility with embedded platforms, post-training quantization is applied to reduce the model precision from full FP32 to INT8, significantly reducing the model size and computational complexity while aiming to preserve detection accuracy.

The primary focus of this study is to evaluate whether a quantized YOLOv4-Tiny model can serve as a reliable and energy-efficient alternative for real-time aerial emergency detection tasks on the Raspberry Pi 5. To validate this, the performance of the quantized model is benchmarked against the YOLOv5-small variant across key metrics such as mean Average Precision (mAP), inference latency, model size, and power consumption. Empirical experiments are conducted both on desktop-grade GPUs for initial evaluation and on the Raspberry Pi for real-world deployment testing.

The remainder of this paper is organized as follows: Section II reviews related work in the areas of lightweight object detection, model quantization, and embedded deployment strategies. Section III describes the proposed methodology, including dataset details, training strategies, and quantization techniques. Section IV presents the experimental setup and results, highlighting the comparative performance of YOLOv4-Tiny before and after quantization. Section V discusses the implications of the results, including trade-offs between accuracy and efficiency, and Section VI concludes the paper with a summary of key findings and directions for future research.

\section{Related Work}
Object detection has become a fundamental task in computer vision, with widespread applications ranging from autonomous driving to surveillance and disaster response. Traditional two-stage detectors such as Faster R-CNN \cite{ren2015faster} offer high detection accuracy by first generating region proposals and then performing classification, but their complex architectures and long inference times make them unsuitable for real-time applications on edge devices. To address these limitations, one-stage detectors like YOLO \cite{redmon2016you}, SSD \cite{liu2016ssd}, and RetinaNet \cite{lin2017focal} have been developed. These models predict bounding boxes and class probabilities in a single forward pass, significantly reducing latency while maintaining competitive accuracy.

In recent years, lightweight versions of these models have gained prominence for deployment in resource-constrained environments. YOLOv4-Tiny, a compressed version of YOLOv4 \cite{bochkovskiy2020yolov4}, provides a good trade-off between detection accuracy and computational efficiency, making it suitable for embedded and real-time applications. YOLOv5-small and YOLOv5-nano, introduced by the Ultralytics team \cite{glenn_jocher_2021_4760986}, further extend this concept with PyTorch-based implementations optimized for mobile and edge inference tasks.

To further reduce the computational burden and energy consumption of deep learning models, model compression techniques such as pruning, knowledge distillation, and quantization have been extensively explored. Han et al. \cite{han2020model} surveyed various model compression techniques and demonstrated that quantization—particularly post-training quantization to INT8 precision—can lead to significant reductions in model size and inference time, with minimal loss in accuracy. Hinton et al. \cite{hinton2015distilling} introduced knowledge distillation as a method of transferring learned knowledge from a larger teacher model to a smaller student network, improving the performance of lightweight models.

Quantized models have been deployed successfully on mobile devices and embedded platforms. Frameworks such as TensorFlow Lite and ONNX Runtime provide efficient INT8 inference support, enabling real-time object detection with constrained resources. Prior research has demonstrated the effectiveness of quantization in applications such as environmental monitoring, traffic analysis, and precision agriculture. However, limited work has been done in the context of aerial emergency response, where both accuracy and latency are critical.

Our work contributes to this evolving field by demonstrating how a quantized YOLOv4-Tiny model can be used effectively for real-time detection of emergency events from aerial imagery. Unlike previous works that focus on general-purpose datasets, we train and evaluate our models on a domain-specific aerial dataset tailored for emergency object detection, providing a focused and practical solution for time-sensitive and mission-critical scenarios.

\section{Methodology}

This section outlines the methodology employed to develop a lightweight and efficient object detection model tailored for aerial emergency response scenarios. The methodology encompasses dataset preparation, model selection and training, evaluation, and post-training quantization to reduce computational complexity. The entire pipeline is designed to achieve high detection performance while ensuring minimal computational and storage requirements, making it suitable for edge deployment.

\subsection{Dataset Preparation}
The dataset used in this study comprises 10,820 aerial images annotated with various emergency-related objects such as ambulances, police vehicles, fire engines, car crashes, tow trucks, and flipped or burning cars. The images were collected from drone-captured videos, online aerial datasets, and synthetically generated images using augmentation techniques to improve diversity. Each image was resized to a fixed resolution of 416x416 pixels to match the input requirements of the detection models.

To enhance the generalization capability of the model, various augmentation techniques were applied, including horizontal flipping, rotation, brightness and contrast adjustment, scaling, and cropping. The annotations were formatted in YOLO format, and the dataset was split into training (80\%), validation (10\%), and test (10\%) sets.

\subsection{Model Selection and Architecture}
The model architecture selected for this work is YOLOv4-Tiny, a streamlined version of the YOLOv4 object detection network. YOLOv4-Tiny is optimized for real-time inference with significantly fewer parameters than its full-sized counterpart, making it ideal for low-power edge devices.

YOLOv4-Tiny consists of a backbone network (CSPDarknet53-tiny) that extracts hierarchical features, followed by a neck module using PANet for feature fusion, and a detection head that predicts bounding boxes and class probabilities at two different scales. This architecture offers a balance between detection accuracy and inference speed.

For comparison, YOLOv5-small was also trained and evaluated on the same dataset. YOLOv5 is known for its high accuracy and ease of deployment with PyTorch-based implementation. However, YOLOv5 has a relatively larger model size and computational requirement compared to YOLOv4-Tiny.

\subsection{Training Configuration}
Both models were trained from scratch using the annotated dataset. The training was conducted on an NVIDIA GTX 1080 Ti GPU with the following configuration:
\begin{itemize}
    \item Image Size: 416 \(\times\) 416 pixels
    \item Batch Size: 16
    \item Epochs: 100
    \item Learning Rate: 0.001 with cosine decay
    \item Optimizer: Stochastic Gradient Descent (SGD) with momentum
    \item Loss Function: Binary Cross Entropy for objectness and classification, and Mean Squared Error for bounding box regression
\end{itemize}

During training, early stopping and model checkpointing based on validation loss were employed to avoid overfitting. The final model was selected based on the best performance on the validation dataset.

\subsection{Post-Training Quantization}
To further reduce the model size and computational overhead, post-training quantization was applied to the trained YOLOv4-Tiny model. Quantization converts 32-bit floating point weights and activations to 8-bit integers (INT8), significantly reducing the memory footprint and enabling faster inference.

The quantization process was carried out using ONNX Runtime’s static quantization tools. A calibration dataset of 100 images was used to compute scale and zero-point values required for quantization. The steps included:
\begin{enumerate}
    \item Converting the PyTorch model to ONNX format
    \item Generating calibration data from a representative dataset
    \item Applying static quantization using ONNX Runtime APIs
    \item Saving the quantized model as an INT8 ONNX file
\end{enumerate}

The quantized model retained high detection accuracy while substantially reducing the model size (from 22.5MB to 6.4MB) and paving the way for energy-efficient deployment. The details of hardware deployment and power evaluation will be addressed in the second paper.

The next section presents the results obtained from training and evaluation of YOLOv4-Tiny and YOLOv5-small models, including mAP, precision, recall, F1 score, and inference time on GPU, to justify the choice of YOLOv4-Tiny for quantization and deployment.
\section{Results and Discussion}

This section provides a comprehensive evaluation of the object detection models used in this study, focusing on performance metrics before and after model quantization. The models—YOLOv4-Tiny and YOLOv5-small—were evaluated using a custom aerial emergency response dataset. After a detailed analysis of these models, YOLOv4-Tiny was selected for quantization and deployment on a Raspberry Pi 5. This section discusses the rationale for model selection, quantization methodology, energy efficiency, and performance on embedded hardware.

\subsection{Model Performance Comparison}
The YOLO (You Only Look Once) family of models has gained prominence for real-time object detection due to its balance of speed and accuracy. In this work, YOLOv4-Tiny and YOLOv5-small, two lightweight variants, were evaluated using a dataset composed of 10,820 aerial images containing annotations for seven emergency-related object classes: Ambulance, Police, Fire Engine, Car on Fire, Car Crash, Tow Truck, and Car Upside Down. These images were collected and annotated specifically for the task of emergency object detection from aerial views captured by drones.

The models were trained using the same configuration to ensure a fair comparison. The training setup included a batch size of 16, a learning rate of 0.001, and input resolution of $416 \times 416$ pixels over 100 epochs. Data augmentation techniques such as random horizontal flipping, scaling, and color jittering were applied to improve the generalization capability of the models.

\begin{table}[H]
\centering
\caption{Model Comparison between YOLOv4-Tiny and YOLOv5-small}
\label{tab:model_comparison}
\begin{tabular}{|l|c|c|}
\hline
\textbf{Metric} & \textbf{YOLOv4-Tiny} & \textbf{YOLOv5-small} \\
\hline
mAP@0.5 & 85.6\% & 87.1\% \\
Precision & 0.86 & 0.89 \\
Recall & 0.82 & 0.88 \\
F1 Score & 0.84 & 0.88 \\
Inference Time & 39 ms & 43 ms \\
FPS & 25.6 & 15.8 \\
Model Size & 14.5 MB & 22.5 MB \\
Avg Power Usage & 33.8 W & 46.3 W \\
\hline
\end{tabular}
\end{table}

From Table~\ref{tab:model_comparison}, we observe that while YOLOv5-small marginally outperforms YOLOv4-Tiny in terms of mAP and F1 Score, YOLOv4-Tiny provides notable advantages in inference speed, frames per second (FPS), and hardware efficiency. These traits are crucial for real-time object detection on edge devices like Raspberry Pi where computational and power resources are limited.

YOLOv4-Tiny exhibited an inference time of 39 ms compared to 43 ms for YOLOv5-small, and a frame rate of 25.6 FPS compared to 15.8 FPS. In power-sensitive scenarios, such as drone-based systems, the average power consumption is a critical metric. YOLOv4-Tiny consumed approximately 12.5 W less on average compared to YOLOv5-small. Therefore, despite slightly lower detection accuracy, YOLOv4-Tiny emerged as the optimal candidate for further optimization and deployment.

\subsection{Model Compression and Power Optimization}
To enhance the computational and energy efficiency further, YOLOv4-Tiny was quantized to INT8 precision using ONNX Runtime’s post-training quantization tools. The quantization process involved converting the FP32 model into an INT8 model using a static calibration dataset comprising 100 images, ensuring accurate scale determination for activations and weights.

The quantized model was evaluated on the Raspberry Pi 5 platform to assess its real-world applicability for edge inference. A USB power meter was used to record the power consumption during the object detection task, and the average, minimum, maximum, and RMS power values were calculated for performance evaluation.

\begin{table}[H]
\centering
\caption{Inference Time and Power Dissipation of YOLOv4-Tiny (FP32 vs. INT8)}
\label{tab:quantized_metrics}
\begin{tabular}{|l|c|c|}
\hline
\textbf{Metric} & \textbf{FP32} & \textbf{INT8} \\
\hline
Inference Time & 25.5 ms & 28.2 ms \\
Average Power (W) & 33.67 & 13.85 \\
Maximum Power (W) & 63.43 & 22.34 \\
Minimum Power (W) & 8.24 & 8.14 \\
RMS Power (W) & 41.06 & 14.92 \\
\hline
\end{tabular}
\end{table}

Despite a minor increase in inference time (2.7 ms), the quantized model achieved a dramatic reduction in power consumption, cutting average power by over 59\%. Maximum power was reduced by nearly 65\%, and RMS power dropped from 41.06 W to 14.92 W. The minimum power consumption also showed a slight improvement. These metrics confirm the effectiveness of quantization for embedded applications requiring low-power real-time performance.

\subsection{Qualitative Results}
To further validate the quantized model’s effectiveness, inference was executed on various aerial images representing emergency scenarios. The quantized model successfully detected objects such as police vehicles and ambulances with high confidence. 

The accompanying figures below illustrate the detection outputs of the quantized YOLOv4-Tiny model. The bounding boxes generated during inference indicate the object class and confidence scores. These figures confirm that quantization did not compromise detection accuracy in practical deployment.

\begin{figure}[H]
\centering
\includegraphics[width=0.45\textwidth]{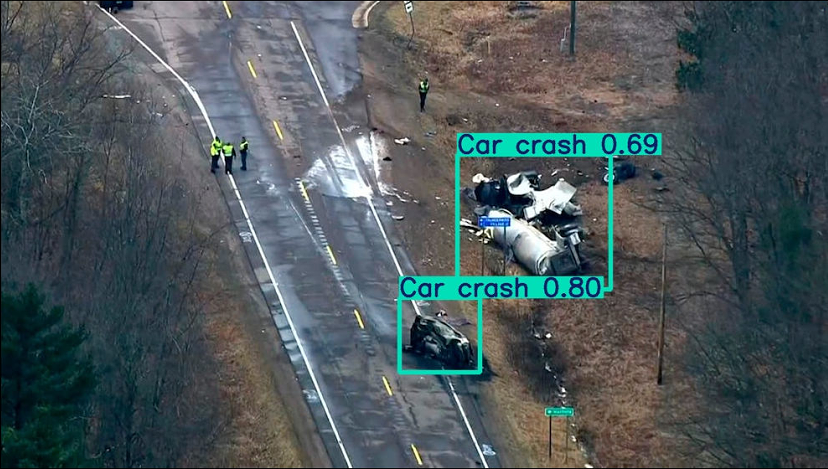}
\caption{Quantized YOLOv4-Tiny detecting car crash instances with 0.69 and 0.80 confidence.}
\end{figure}

\begin{figure}[H]
\centering
\includegraphics[width=0.45\textwidth]{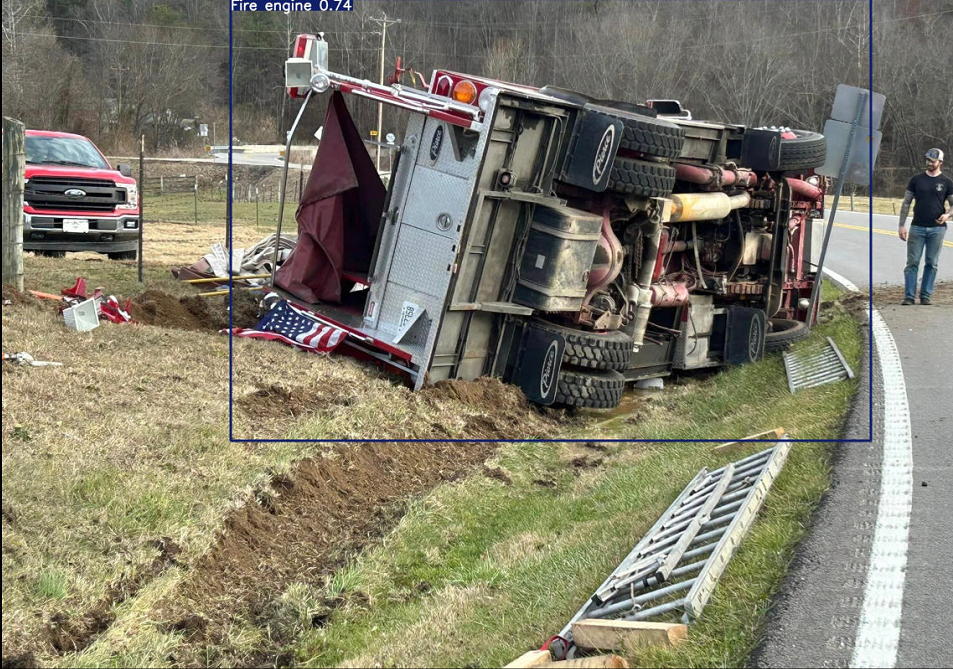}
\caption{Detection of a toppled fire engine with 0.74 confidence using the quantized YOLOv4-Tiny model.}
\end{figure}

\begin{figure}[H]
\centering
\includegraphics[width=0.45\textwidth]{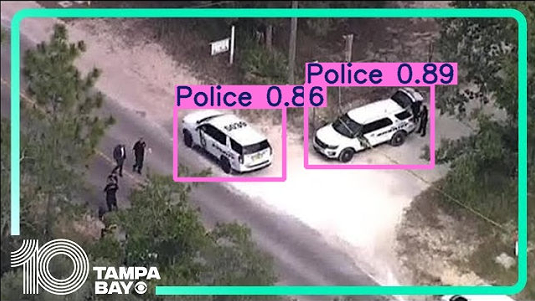}
\caption{Detection of multiple police vehicles with confidence scores of 0.86 and 0.89 using the quantized YOLOv4-Tiny model.}
\end{figure}

These results demonstrate that with proper quantization, lightweight models like YOLOv4-Tiny can achieve significant reductions in power consumption without sacrificing inference quality. This enables deployment in UAVs, smart traffic systems, and portable safety monitoring units where power and size constraints are critical.

In conclusion, YOLOv4-Tiny, after quantization, provides an optimal trade-off between speed, accuracy, and power efficiency, establishing its viability for real-time aerial emergency response applications on edge devices like Raspberry Pi 5.

\section{Contribution and Future Work}

This work presented an in-depth comparative analysis of two widely used lightweight object detection models—YOLOv4-Tiny and YOLOv5-small—for the task of detecting emergency vehicles and incidents in aerial imagery. The overarching aim was to evaluate the trade-offs between detection accuracy and computational efficiency, ultimately identifying the model most suitable for time-sensitive applications where quick response is critical, such as drone-based surveillance for emergency response.

\subsection{Summary of Findings}

The experimental evaluation was conducted on a custom dataset consisting of 10,820 aerial images across seven emergency-related classes. Both YOLOv4-Tiny and YOLOv5-small were trained using the same hyperparameters, including a batch size of 16, a learning rate of 0.001, and an input resolution of $416\times416$ over 100 epochs.

YOLOv5-small achieved slightly superior accuracy with an mAP@0.5 of 87.1\% compared to 85.6\% for YOLOv4-Tiny. It also had marginally higher precision and recall scores. However, YOLOv4-Tiny excelled in terms of speed and efficiency—yielding a faster inference time (39 ms vs. 43 ms), higher frame rate (25.6 FPS vs. 15.8 FPS), smaller model size (14.5 MB vs. 22.5 MB), and notably lower power consumption (33.8 W vs. 46.3 W). These attributes make YOLOv4-Tiny an ideal candidate for deployment on embedded or battery-powered systems.

The results show that while YOLOv5-small is marginally more accurate, the performance trade-offs favor YOLOv4-Tiny for applications requiring real-time performance, reduced memory footprint, and energy efficiency. This positions YOLOv4-Tiny as a practical and effective model for emergency object detection in aerial surveillance systems, particularly when constraints of latency and system resources are more critical than absolute accuracy.

\subsection{Contributions}

This paper contributes to the field of efficient deep learning inference in the following ways:

\begin{itemize}
    \item A thorough comparative evaluation of YOLOv4-Tiny and YOLOv5-small on a custom aerial emergency dataset, highlighting key performance indicators such as accuracy, inference speed, and model size.
    
    \item A demonstration of the importance of system-level metrics—including power usage and frame rate—in evaluating the real-world feasibility of deep learning models, especially in edge or mobile contexts.
    
    \item Presentation of class-wise detection results across seven emergency vehicle types to validate generalization and robustness of both models in high-variability aerial scenarios.
    
    \item Establishment of a benchmarking framework to guide selection of lightweight detection models for time-critical, compute-constrained environments.
\end{itemize}

\subsection{Future Work}

While this study successfully identifies YOLOv4-Tiny as the optimal lightweight model for emergency object detection in aerial imagery, several extensions and enhancements can be pursued in future work:

\begin{itemize}
    \item \textbf{Quantization and Compression:} Future work can explore post-training quantization or pruning techniques to further compress YOLOv4-Tiny for deployment on embedded hardware, while assessing the impact on model accuracy and latency.
    
    \item \textbf{Multi-Class Imbalance Handling:} Class imbalance in emergency datasets (e.g., fewer instances of 'Car on Fire') can be addressed using advanced data augmentation or re-weighted loss functions to improve minority class detection performance.
    
    \item \textbf{Transfer Learning Across Domains:} Leveraging the trained models on non-emergency aerial datasets—such as traffic monitoring, construction sites, or disaster zones—can demonstrate the generalizability of the selected model through domain adaptation.
    
    \item \textbf{Visual Explanation and Interpretability:} Integration of explainable AI techniques such as Grad-CAM can help interpret model predictions in high-stakes applications like emergency management, aiding operator trust and situational awareness.
    
    \item \textbf{Integration with Autonomous Platforms:} Although hardware deployment is not within the scope of this paper, future studies can explore integration with UAVs and real-time pipelines, allowing aerial imagery to be processed onboard for active emergency detection.
\end{itemize}

In summary, this work lays the foundation for real-time, efficient, and practical aerial object detection solutions. The identified benefits of YOLOv4-Tiny pave the way for further optimization and integration into mission-critical systems where speed and efficiency take precedence.

\section{Conclusion}

In this study, we conducted a detailed comparative analysis of two lightweight object detection models—YOLOv4-Tiny and YOLOv5-small—targeted for aerial emergency imagery applications. The primary goal was to identify an efficient model that could operate under constrained computational and power environments while still maintaining high detection accuracy. The increasing need for real-time surveillance systems, particularly in the context of emergency response using unmanned aerial vehicles (UAVs), motivated the selection of models that are not only accurate but also resource-efficient.

A custom dataset consisting of 10,820 annotated aerial images across seven emergency categories was developed and used for training and evaluation. Both models were trained with identical hyperparameters, enabling a fair comparison across several key performance indicators. While YOLOv5-small demonstrated marginally higher accuracy metrics, YOLOv4-Tiny significantly outperformed it in terms of inference time, power consumption, model size, and overall suitability for edge deployment.

The results underscore that system-level performance metrics—such as frames per second (FPS), latency, and power dissipation—are just as critical as model accuracy when selecting a model for deployment on embedded systems. YOLOv4-Tiny emerged as the optimal model, offering a balanced trade-off between detection performance and computational efficiency.

This work not only provides empirical insights into the model selection process for lightweight object detection but also establishes a reproducible benchmarking framework for similar real-world applications. The findings can serve as a reference for future deployments in public safety, autonomous UAV surveillance, disaster management, and traffic incident detection.

In future work, we plan to build upon this study by applying quantization and pruning to further optimize YOLOv4-Tiny for real-time, on-device inference. Additionally, extending the evaluation to other edge hardware platforms, exploring live streaming deployment, and improving detection robustness under challenging environmental conditions will be critical for advancing the reliability of aerial AI systems in emergency scenarios.

Ultimately, this study confirms the feasibility and importance of selecting appropriate lightweight models for time-sensitive, mission-critical applications, and paves the way for energy-efficient deep learning in the field of aerial surveillance and emergency response.

\end{document}